\documentclass[letterpaper, 10 pt, conference]{ieeeconf}  

\IEEEoverridecommandlockouts                              

\usepackage{graphicx} 
\usepackage{amsmath} 
\usepackage{amssymb}  
\usepackage{multirow}
\usepackage{colortbl}
\usepackage{subfig}

\title{\LARGE \bf
In Situ Calibration of Six-Axis Force-Torque Sensors \\ using Accelerometer Measurements*
}

\author{Silvio Traversaro and Daniele Pucci and Francesco Nori
\thanks{*This paper was supported by the FP7 EU projects CoDyCo (No. 600716
ICT 2011.2.1 Cognitive Systems and Robotics) and Koroibot (No. 611909
ICT-2013.2.1 Cognitive Systems and Robotics).}
\thanks{All authors belong to the Istituto Italiano di Tecnologia, RBCS department, Genova, Italy. Email: {\tt\small name.surname@iit.it}}%
}

\begin{document}

\newcommand{\prop}[2]{\textbf{Proposition #1. }\textit{#2}}
\newcommand{\remark}[2]{\textbf{Remark #1. }\textit{#2}}
\newcommand{\lemma}[2]{\textbf{Lemma #1. }\textit{#2}}
\newcommand{\hp}[2]{\textbf{Assumption #1. }\textit{#2}}
\newcommand{\nrofsg}{6}
\newcommand{\calibmat}{\mathbf{C}}
\newcommand{\shapemat}{\mathbf{S}}
\newcommand{\rawval}{\mathbf{r}}
\newcommand{\senswrench}{{}^s\mathbf{w}}
\newcommand{\sensgrav}{{}^s\mathbf{g}}
\newcommand{\offsetwrench}{{}^\mathbf{o_w}}
\newcommand{\offsetraw}{\mathbf{o_r}}

\maketitle
\thispagestyle{empty}
\pagestyle{empty}

\newtheorem{hypothesis}{Assumption}

\begin{abstract}
This paper proposes techniques to calibrate six-axis force-torque sensors that can be performed \emph{in situ}, i.e., without removing the sensor from
the hosting 
system. 
We assume that the force-torque sensor is attached to a rigid body equipped with an accelerometer. Then, the proposed calibration 
technique uses the measurements of the accelerometer, but
requires neither the knowledge of the inertial parameters 
nor the orientation
of the rigid body. 
The proposed method exploits the geometry induced by the model between the raw measurements of the 
sensor and the corresponding force-torque. 
The validation of the approach is performed by calibrating two six-axis force-torque sensors of the iCub humanoid 
robot.
\end{abstract}

\section{INTRODUCTION}
The importance of sensors in a control loop goes without saying.
Measurement devices, however, can seldom be used \emph{sine die} without being subject to periodic calibration procedures.
Pressure transducers, gas sensors, and thermistors are only a few examples of measuring devices that need to be calibrated periodically for providing 
the user with precise and robust measurements.
Of most importance, calibration procedures may require to move the sensor from the hosting system to specialized laboratories, which are equipped with
the tools for performing the calibration of the measuring device. This paper presents techniques to calibrate 
strain gauges six-axis force-torque sensors \emph{in situ}, i.e. without the need of removing the sensor from the hosting system.
The proposed calibration method is particularly suited for robots with embedded six-axis force-torque sensors installed within limbs~\cite{Fumagalli2012}. 

Calibration of six-axis force-torque sensors  has long attracted the attention of the robotic community~\cite{braun2011}. 
The commonly used model for predicting the force-torque 
from the raw measurements of the sensor is an affine model.
This model is sufficiently accurate 
since these sensors are mechanically designed and mounted so that the strain deformation is (locally) linear with respect to the applied forces and torques.
Then, the calibration of the sensor aims at determining the two components of this model, i.e. a six-by-six matrix and a six element vector.
These two components are usually referred to as the sensor's \emph{calibration matrix} and \emph{offset}, respectively.
In standard operating conditions, relevant changes in the calibration matrix may occur in months.
As a matter of fact, manufacturers such as ATI~\cite{atimanual} and Weiss Robotics~\cite{kms40manual} 
recommend to calibrate force-torque sensors at least once a year.
Preponderant changes in the sensor's offset can occur in hours, however, and this in general requires to estimate the 
offset before using the  sensor.
The offset models the sensitivity of the semiconductor strain gauges with respect to temperature.

\begin{figure}
\vspace{0em}
\centering
\includegraphics[width=0.225\textwidth]{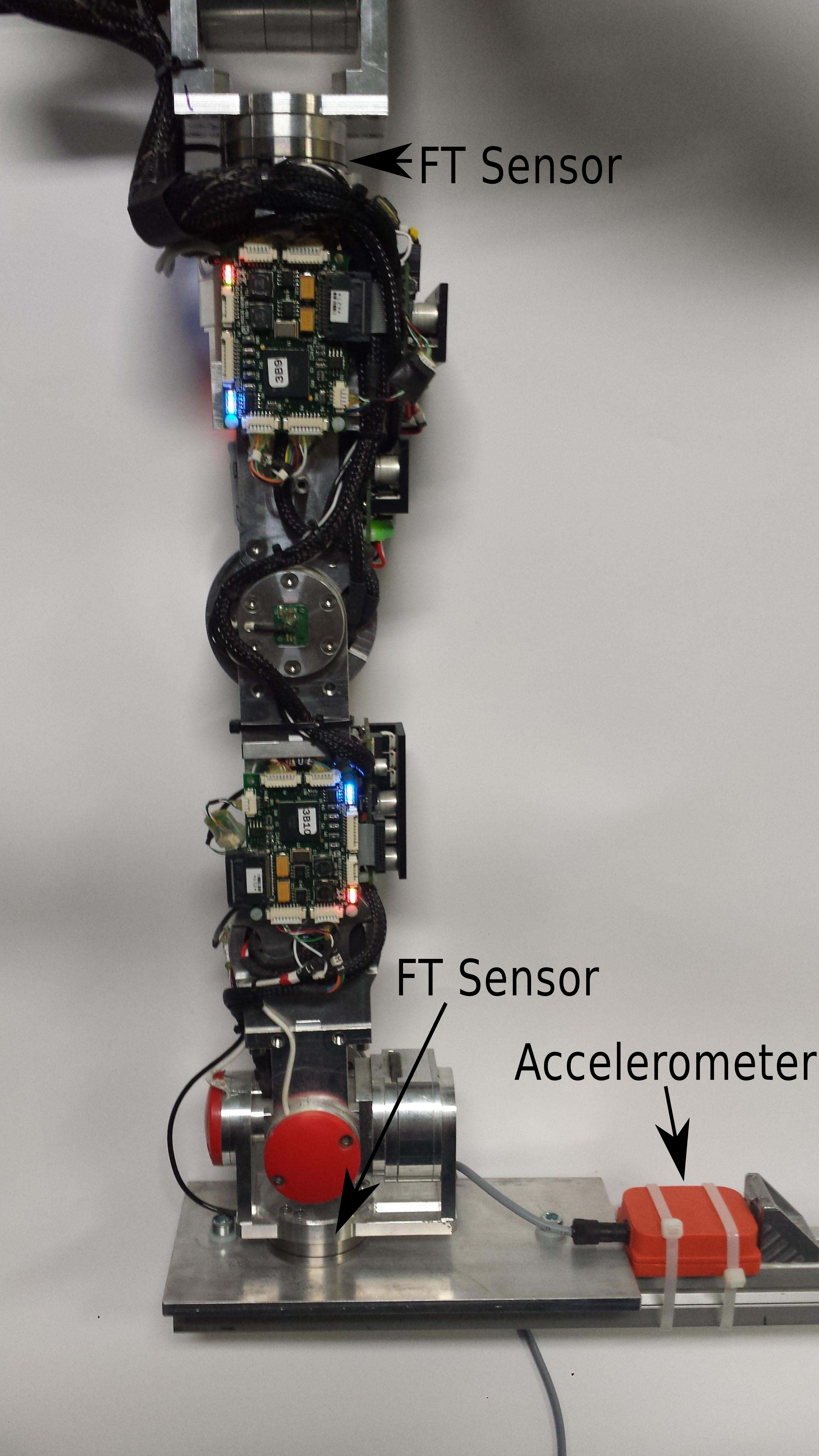};
\caption{iCub's leg with the two force/torque sensors and an additional accelerometer.}
\label{fig:iCubLeg}
\end{figure} 
Classical techniques for determining the offset of a force-torque sensor exploit the aforementioned affine model between 
the raw measurements and the load attached to the sensor. In particular, if no load is applied to the measuring device, the output of the sensor corresponds to 
the sensor's offset. This offset identification procedure, however, cannot be always performed since it may require to take the hosting system apart
in order to unload the force-torque sensor. Another widely used technique for offset identification is to find two sensor's orientations that induce
equal and opposite loads with respect to the sensor. Then, by summing up the raw measurements associated with these two orientations, one can estimate
the sensor's offset. The main drawback of this technique is that the positioning of the sensor at these opposite configurations may require to move
the hosting system beyond its operating domain.

Assuming a preidentified offset, non-in situ identification of the calibration matrix is classically performed by exerting on  
the sensor a set of force-torques known \emph{a priori}. This usually requires to place 
sample masses at precise relative positions with respect to the sensor. Then, by comparing
the known gravitational force-torque with that measured by the sensor, 
one can apply linear least square techniques to identify the sensor's calibration matrix.
For accurate positioning of the sample masses, the use of robotic positioning devices 
has also been proposed in the specialized 
literature~\cite{uchiyama1991systematic}~\cite{watson1975pedestal}.

To reduce the number of sample masses,
one can apply constrained forces, e.g. constant norm forces, to the measuring device.
Then these constrains can be exploited during the computations for identifying the calibration matrix~\cite{voyles1997shape}.
To avoid the use of added masses, one can use a supplementary already-calibrated measuring device that measures 
the force-torque exerted on the sensors~\cite{faber2012force}~\cite{oddo2007}.
On one hand, this calibration technique avoids the problem of precise positioning of the added sample masses.
On the other hand, 
the supplementary sensor may not always be available.
All above techniques, however, cannot be performed in situ, thus  being usually time consuming and expensive.

To the best of our knowledge, the first \emph{in situ} calibration method for force-torque sensors was proposed 
in \cite{shimanoroth}. But this method  exploits the topology of a specific kind of manipulators, which are equipped with
joint torque sensors then leveraged during the estimation. 
A recent result~\cite{Gong2013} proposes another \emph{in situ} calibration technique for six-axis force-torque sensors. 
The technical soundness of this work, however, is not clear to us. In fact, we show that a necessary condition for identifying the calibration matrix
is to take measurements for at least three different added masses, and this requirement was not met by the algorithm~\cite{Gong2013}.
Another in situ calibration technique for force-torque sensors can be found in \cite{roozbahani2013novel}.
But the use of supplementary already-calibrated force-torque/pressure sensors impairs this technique for the reasons we have discussed before.

This paper presents in situ calibration techniques for six-axis force-torque sensors using accelerometer measurements.
The proposed method exploits the geometry induced by the affine model between the raw measurements and the gravitational force-torque applied to the sensor. 
In particular, it relies upon the properties that all gravitational raw measurements belong to a three-dimensional space, and that in this space they form an ellipsoid.
Then, the contribution of this paper is twofold. We first propose a method for estimating the sensor's offset, and then a procedure for identifying
the calibration matrix. The latter is independent from the former, but requires to add sample masses to the rigid body attached to the sensor. Both methods are independent from the inertial characteristics 
of the rigid body attached to the sensor. 
The proposed algorithms are validated on the iCub platform by calibrating
two force-torque sensors embedded in the robot~leg.

The paper is organized as follows. Section~\ref{sec:background} presents the notation used in the paper and the problem statement with the assumptions. 
Section~\ref{method} details the proposed method for the calibration of six-axis force-torque sensors. 
Validations of the approach are presented in Section~\ref{experiments}.
Remarks and perspectives conclude the paper.

\section{BACKGROUND}
\label{sec:background}

\subsection{Notation}
The following notation is used throughout the paper.
\begin{itemize}
 \item The set of real numbers is denoted by $\mathbb{R}$. Let $u$ and $v$ be two $n$-dimensional column vectors of real numbers, i.e. $u,v \in \mathbb{R}^n$, 
 their inner product is denoted as $u^\top v$, with ``$\top$'' the transpose operator.
\item Given $u \in \mathbb{R}^3$, $u \times$ denotes the skew-symmetric matrix-valued operator associated with the cross product in 
  $\mathbb{R}^3$.
 \item The Euclidean norm of either a vector or a matrix of real numbers is denoted by $|\cdot |$.
\item $I_n \in \mathbb{R}^{n \times n}$ denotes the identity matrix of dimension~$n$; 
$0_n \in \mathbb{R}^n$ denotes the zero column vector of dimension~$n$; $0_{n \times m} \in \mathbb{R}^{n \times m}$ denotes the zero matrix of dimension~$n \times m$.
\item The vectors $e_1,e_2,e_3$ denote the canonical basis of $\mathbb{R}^3$.
\item Let $\mathcal{I} = \{O;e_1,e_2,e_3\}$ denote a 
fixed inertial frame with respect to (w.r.t.) which the sensor's absolute orientation is measured. 
Let $\mathcal{S} = \{O';i,j,k\}$ denote a frame attached to the sensor, where the matrix $T := (i,j,k)$ is a rotation matrix
whose column vectors are the vectors
of coordinates of $i,j,k$ expressed in the basis of  $\mathcal{I}$. 
\item The sensor's orientation w.r.t. $\mathcal{I}$ is characterized by the rotation matrix $T$. Given a vector of coordinates $\bar{x} \in \mathbb{R}^3$
expressed w.r.t. $\mathcal{I}$, we denote with $x$ the same vector expressed w.r.t. $\mathcal{S}$, i.e. $\bar{x} = Tx$.
\item Given $A \in \mathbb{R}^{n \times m}$ and $B \in \mathbb{R}^{p \times q}$, we denote with $\otimes$ the Kronecker product $A \otimes B \in \mathbb{R}^{np \times mq}$.
\item Given $X \in \mathbb{R}^{m \times p}$, $\text{vec}(X) \in \mathbb{R}^{nm}$ denotes the column vector obtained by stacking the columns of the matrix~$X$. 
In view of the definition of $\text{vec}(\cdot)$, it follows that \begin{equation}\label{eq:kroneckerVec} \text{vec}(AXB) = \left( B^{\top} \otimes A \right) \text{vec}(X).\end{equation}

\end{itemize}

\subsection{Problem statement and assumptions}
We assume that the model for predicting the force-torque (also called wrench)  
from the raw measurements is an affine model, i.e. 
\begin{equation}
\label{wrenchInSensorCoordinates}
w =  C ( r - o),
\end{equation} 
where
$ {w} \in \mathbb{R}^{6}$ is the wrench exerted on the sensor expressed in the sensor's frame,
$r \in \mathbb{R}^{\nrofsg}$ is the raw output of the sensor,
$ {C} \in \mathbb{R}^{6 \times \nrofsg}$ is the
invertible
calibration matrix, and
${o} \in \mathbb{R}^6$ is the sensor's offset.
The calibration matrix and the offset are assumed to be constant.

We assume that the sensor is attached to a rigid body
of 
(constant) mass $m~\in~\mathbb{R}^+$ and with a center of mass whose position
w.r.t. the sensor frame $\mathcal{S}$ is characterized the vector $c \in \mathbb{R}^3$.

The gravity force applied to the 
body
is  given by 
\begin{IEEEeqnarray}{RCL}
 \label{eq:g}
 m\bar{g} = mTg ,
\end{IEEEeqnarray}
with $\bar{g},g \in \mathbb{R}^3$ the gravity acceleration expressed w.r.t. the inertial and sensor frame, respectively. The gravity acceleration $\bar{g}$ is constant, so the vector $g$ does not have a constant direction, 
but has a constant norm.

Finally, we make the following main assumption.

\hp{}{
 The raw measurements $r$ are taken for static configurations of the 
 rigid body
 attached to the sensor, i.e. the angular velocity of the frame $\mathcal{S}$ is always zero.
 Also, the gravity acceleration $g$ is measured by an accelerometer installed on the rigid body.
  Furthermore, no external force-torque, but the gravity force, acts on the rigid body. Hence
}
\begin{IEEEeqnarray}{RCL}
\label{eq:staticWrench}
w &=& 
M(m,c) g, \IEEEyessubnumber  \\
M(m,c) &:=& m
\begin{pmatrix}
I_3 \\
c \times 
\end{pmatrix}. \label{matrixM}  \IEEEyessubnumber 
\end{IEEEeqnarray} 

\remark{}{ We implicitly assume that the accelerometer frame is aligned with the force-torque sensor frame. 
This is a convenient, but non necessary, assumption. 
In fact, if the relative rotation between the sensor frame $\mathcal{S}$ and the accelerometer frame is unknown, 
it suffices to consider the accelerometer frame as the sensor frame $\mathcal{S}$.
}

Under the above assumptions, what follows proposes a new method for estimating the sensor's offset $o$ 
and for identifying the sensor's calibration matrix $C$ without the need of removing the sensor from the 
hosting system.

\section{METHOD}
\label{method}
The proposed methods rely on the the geometry induced by the models~\eqref{eq:staticWrench} and~\eqref{wrenchInSensorCoordinates}.

\subsection{The geometry of the raw measurements}
\label{subsec:geometry}
First, observe that
\[ \text{rank}(M) = 3 .\]
As a consequence, all wrenches $w$ belong to the three-dimensional subspace given by the $\operatorname{span}{(M)} \subset \mathbb{R}^6$. In view of this,
we can state the following lemma.

\lemma{1}{All raw measurements $r$ belong to a three dimensional affine space,  i.e. there exist a point $r_m \in \mathbb{R}^6$, an orthonormal basis 
$U_1 \in \mathbb{R}^{6 \times 3}$, and for each $r \in \mathbb{R}^6$ a vector $\lambda \in \mathbb{R}^3$ such that 
\begin{IEEEeqnarray}{RCL}
 \label{decompositionMeasurements}
 r &=& r_m + U_1 \lambda.
\end{IEEEeqnarray}
Also, the vector $\lambda$ belongs to a three-dimensional ellipsoid.
}
\begin{proof}
From \eqref{wrenchInSensorCoordinates} and \eqref{eq:staticWrench}, one has:
\begin{equation}
\label{rFromModel}
r  = |g|C^{-1} M \hat{g} + o,
\end{equation}
where $\hat{g} := g/|g|$.
The matrix $C^{-1}M \in \mathbb{R}^{6\times3}$ is of rank~$3$.
Consequently, all raw measurements $r$ belong to an affine three-dimensional space 
defined by the point $o$ and the basis of $\operatorname{span}{(C^{-1}M)}$.
Now, define $P \in \mathbb{R}^{3 \times 6}$ as the projector of $r$ onto $\operatorname{span}{(C^{-1}M)}$.
 Then, the projection $p \in \mathbb{R}^3$ of $r$ onto this span is given by  
\begin{IEEEeqnarray}{RCL}
  p &=& Pr = |g|PC^{-1} M \hat{g} + Po.
\end{IEEEeqnarray} 
By considering all possible orientations of the sensor's frame~$\mathcal{S}$, then the gravity direction $\hat{g}$ spans the unit sphere. 
Consequently, the vector $p$ belongs to the \textit{span of the unit sphere applied to the linear transformation $|g|PC^{-1} M$}, 
i.e. an \textit{ellipsoid centered at the point $Po$}. This in turn implies that when decomposing the vector $r$ as in~\eqref{decompositionMeasurements},
the vector~$\lambda$ necessarily belongs to a three-dimensional ellipsoid.
\end{proof}

\begin{figure}
\vspace{2em}
\centering
    \includegraphics[width=0.45\textwidth]{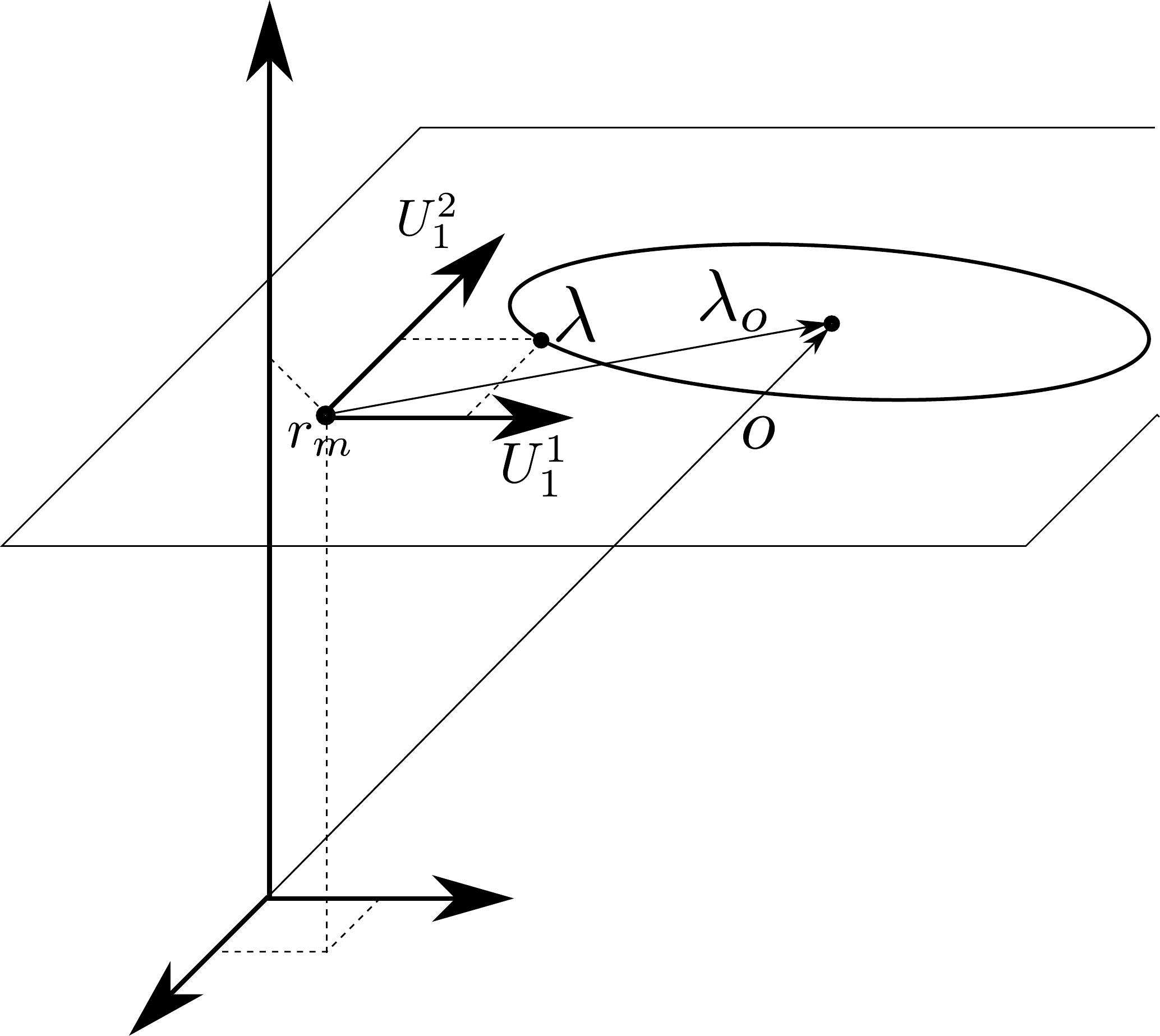}
    \caption{Example when $r \in \mathbb{R}^3$ and $U_1 = (U_1^1,U_1^2) \in \mathbb{R}^{3 \times 2}$. }
    \label{imageWithPlane}
\end{figure}

To provide the reader with a better comprehension of the above lemma, assume that $r \in \mathbb{R}^{3}$ and that the affine subspace is 
a plane, i.e. a two-dimensional space. 
As a consequence, all measurements belong to an ellipse lying on this plane -- see Figure~\ref{imageWithPlane}. Observe also that given a point 
$\lambda \in \mathbb{R}^2$ on the plane and expressed w.r.t. the basis $U_1$, the relationship~\eqref{decompositionMeasurements} provides with the components of this point in the space 
$\mathbb{R}^3$.
By leveraging on the above lemma, the next two sections propose a method to estimate the sensor's offset and calibration matrix.

\subsection{Method for estimating the sensor's offset}
\label{offsetEstimationTechnique}

Assume that one is given with a set of measurements $(r_i,g_i)$ with $i \in \{1 \cdots N \}$ corresponding to several body's static orientations. 
Then, let 
us show how we can obtain the basis $(r_m,U_1)$ in~\eqref{decompositionMeasurements} associated with all measurements $r_i$, and how this basis
can be used for estimating the offset $o$.

Observe that the point $r_m$ can be chosen as any point that belongs to the affine space. Then, 
a noise-robust choice for this point is given by the mean value of the measurements~$r_i$,
\begin{IEEEeqnarray}{RCL}
 r_m &=& \frac{1}{N} \sum\limits_{i=1}^N r_i.
\end{IEEEeqnarray}
An orthonormal basis $U_1$ can be then obtained by applying the singular value decomposition on the matrix resulting from the difference between  
all measurements and $r_m$, i.e.
\begin{IEEEeqnarray}{RCL}
 (\tilde{r}_1, \cdots, \tilde{r}_n) = USV^\top,  
\end{IEEEeqnarray}
where 
\[\tilde{r}_i := r_i - r_m,\] and $U \in \mathbb{R}^{6\times6}$, $S \in \mathbb{R}^{ 6 \times N}$, $V \in \mathbb{R}^{N\times N}$ are 
the (classical) matrices obtained from the singular value decomposition. 
Note that
only the first three elements on the diagonal of$~S$ are (significantly) different from zero
since all measurements~$r_i$ belong to a three dimensional subspace. Consequently, (an estimate of) the orthonormal basis $U_1$ is given by the first
three columns of the matrix~$U$. 
 
With $(r_m,U_1)$ in hand, the offset $o$ can be easily estimated.
First, note that equation \eqref{decompositionMeasurements} holds for all points belonging to the three dimensional space.
Hence, it holds also for the offset $o$ being the center of the ellipsoid (see Figure~\ref{imageWithPlane}), i.e. 
\begin{equation}
\label{eq:offsetInPlane}
o = r_m + U_1 \lambda_o .
\end{equation}
Then to estimate the offset $o$ belonging to $\mathbb{R}^6$, we can estimate the coordinates $\lambda_o$ in the subspace $\mathbb{R}^3$. In view of $U_1^\top U_1 = I_3$, multiplying
the above equation times $U_1^\top$ yields
\begin{equation}
\lambda_o := U^{\top}_1(o-r_m).
\end{equation}
Now, by subtracting $r_m$ from \eqref{rFromModel} and multiplying the resulting equation by $U_1^\top$, one has
\begin{equation}
 \label{rFromModel1}
 U^{\top}_1\tilde{r}_i  = Kg_i + \lambda_o,
\end{equation}
where 
\[K := U^\top_1C^{-1}M \quad \text{and} \quad \lambda_o \] are the unknowns
in the above equation. In view of~\eqref{eq:kroneckerVec} the equation~\eqref{rFromModel1} can be written by stacking the obtained vectors for all measurements as
\begin{IEEEeqnarray}{RCLRCL}
 \bar{r}  &=& \Gamma x,  \IEEEyessubnumber \label{eqForOffset} \\
\bar{r} &:=& 
 \begin{pmatrix}
  \tilde{r}^\top_1U_1, && \cdots, && \tilde{r}^\top_NU_1
 \end{pmatrix}^\top &\in& \mathbb{R}^{3N\times 1},  \IEEEyessubnumber \\ 
 \Gamma &:=& 
 \begin{pmatrix}
  g^\top_1 \otimes I_3, I_3 \\
  . \\
  . \\
  g^\top_N \otimes I_3, I_3
 \end{pmatrix}&\in& \mathbb{R}^{3N\times 12},  \IEEEyessubnumber \\
 x &:=& 
 \begin{pmatrix}
  \text{vec}(K) \\
  \lambda_o
 \end{pmatrix}&\in& \mathbb{R}^{12\times 1}.\IEEEyessubnumber
\end{IEEEeqnarray}
Solving the equation~\eqref{eqForOffset} for the unknown $x$ in the least-square sense provides with an estimate 
$\hat{\lambda}_o  \in \mathbb{R}^3$ of $\lambda_o$. To obtain the coordinates of this point w.r.t. the six-dimensional space, i.e. the raw measurements space,
we apply the transformation~\eqref{eq:offsetInPlane} as follows:
\[
\hat{o} = r_m + U_1 \hat{\lambda}_o .
\]

\subsection{Method for estimating the  sensor's calibration matrix}
\label{calibrationMatrixEstimation}

In this section, we assume no offset, i.e. $o = 0$, which means that this offset has already been estimated
by using one of the existing methods in the literature 
or by using the method described in the previous section. Consequently, the relationship between the set of measurements 
$(r_i,g_i)$ 
and the body's inertial characteristics, i.e. mass and center of mass, is given by
\begin{equation}
Cr_i =  Mg_i \nonumber.
\end{equation} 
In addition, we also assume that the body's inertial characteristics can be modified by adding sample masses at specific relative positions w.r.t. the
sensor frame $\mathcal{S}$. As a consequence, the matrix $M$ in the above equation is replaced by $M_j$, i.e.
\begin{equation}
\label{rawMeasurementsNoOffsetSeveralDataSets}
Cr^j_i =  M_jg^j_i ,
\end{equation} 
where $j$ indicates that new inertial
characteristics have been obtained by adding sample masses. Observe that $M_j$ can then be decomposed as follows
\begin{IEEEeqnarray}{RCCRCL}
 \label{MknownUnkwn}
 M_j &:= &M_b& + &M^j_a& \nonumber \\
    &=& m
\begin{pmatrix}
I_3 \\
c \times 
\end{pmatrix}&
+ &m^j_a
\begin{pmatrix}
I_3 \\
c^j_a \times 
\end{pmatrix}&,  \nonumber
\end{IEEEeqnarray}
where $(m^j_a,c^j_a)$ are the mass and the vector of the center of mass, expressed w.r.t. the sensor frame $\mathcal{S}$, of the added mass.
In the above equation, $M_b$ is unknown but $M^j_a$ is assumed to be known.

In light of the above, we assume to be given with several data sets 
\begin{IEEEeqnarray}{RCL}
 \label{dataSets}
 R_j := (r^j_1, \cdots,r^j_{N_j}) \in \mathbb{R}^{6\times N_j}, \IEEEyessubnumber \\
 G_j := (g^j_1, \cdots,g^j_{N_j}) \in \mathbb{R}^{3\times N_j} ,\IEEEyessubnumber 
\end{IEEEeqnarray}
associated with $N_D$ different $(m^j_a,c^j_a)$. Given~\eqref{rawMeasurementsNoOffsetSeveralDataSets} and~\eqref{dataSets},
the measurements associated with the $jth$ data set can be compactly written as
\begin{IEEEeqnarray}{RCL}
CR_j - M_bG_j&=&   M^j_aG_j. \nonumber
\end{IEEEeqnarray} 
The matrices $C$ and $M_b$ are unknown. Then, in view of~\eqref{eq:kroneckerVec} the above equation can be written for all data sets as follows
\begin{IEEEeqnarray}{RCLRLL}
 \Theta x &=& \beta,  \IEEEyessubnumber \label{equationForEstimatingC} \\
 x &:=& 
 \begin{pmatrix}
  \text{vec}(C) \\
  m \\
  mc
 \end{pmatrix},
 &\in& \mathbb{R}^{40\times 1},  
 \IEEEyessubnumber \\ 
 \Theta &:=& 
 \begin{pmatrix}
  R^\top_1 \otimes I_6, \ -(G^\top_1 \otimes I_6)H  \\
  . \\
  . \\
  R^\top_{N_D} \otimes I_6, \ -(G^\top_{N_j} \otimes I_6)H
 \end{pmatrix},
 &\in& \mathbb{R}^{6N_T\times 40},
 \IEEEyessubnumber \IEEEeqnarraynumspace \\
 \beta &:=& 
 \begin{pmatrix}
  \text{vec}(M^1_a G_1)   \\
  . \\
  . \\
  \text{vec}(M^{N_D}_a G_1) 
 \end{pmatrix},
 &\in& \mathbb{R}^{40\times 1}.
 \IEEEyessubnumber
\end{IEEEeqnarray}
with \[N_T = \sum\limits_{j=1}^{N_D} N_j,\] i.e. the number of all measurements, and the matrix $H~\in~\mathbb{R}^{18\times4}$ a properly chosen permutator such that 
\[\text{vec}(M_b) = H
 \begin{pmatrix}
  m \\
  mc
 \end{pmatrix}.\]
To find the calibration matrix $C$, we have to find the solution~$x$ to the equation~\eqref{equationForEstimatingC}. 
The uniqueness of this solution is characterized by the following lemma.

\lemma{2}{
A necessary condition for the uniqueness of 
the solution $x$ to the equation~\eqref{equationForEstimatingC} is that the number of data sets 
must be greater 
than two, i.e.
\begin{IEEEeqnarray}{rcl}
 \label{necessityForWellPosedenessC}
 N_D \geq 3.
\end{IEEEeqnarray}
}

\begin{figure}[t]
\centering
\subfloat[Dataset 1: no added mass]{\includegraphics[width=0.22\textwidth]{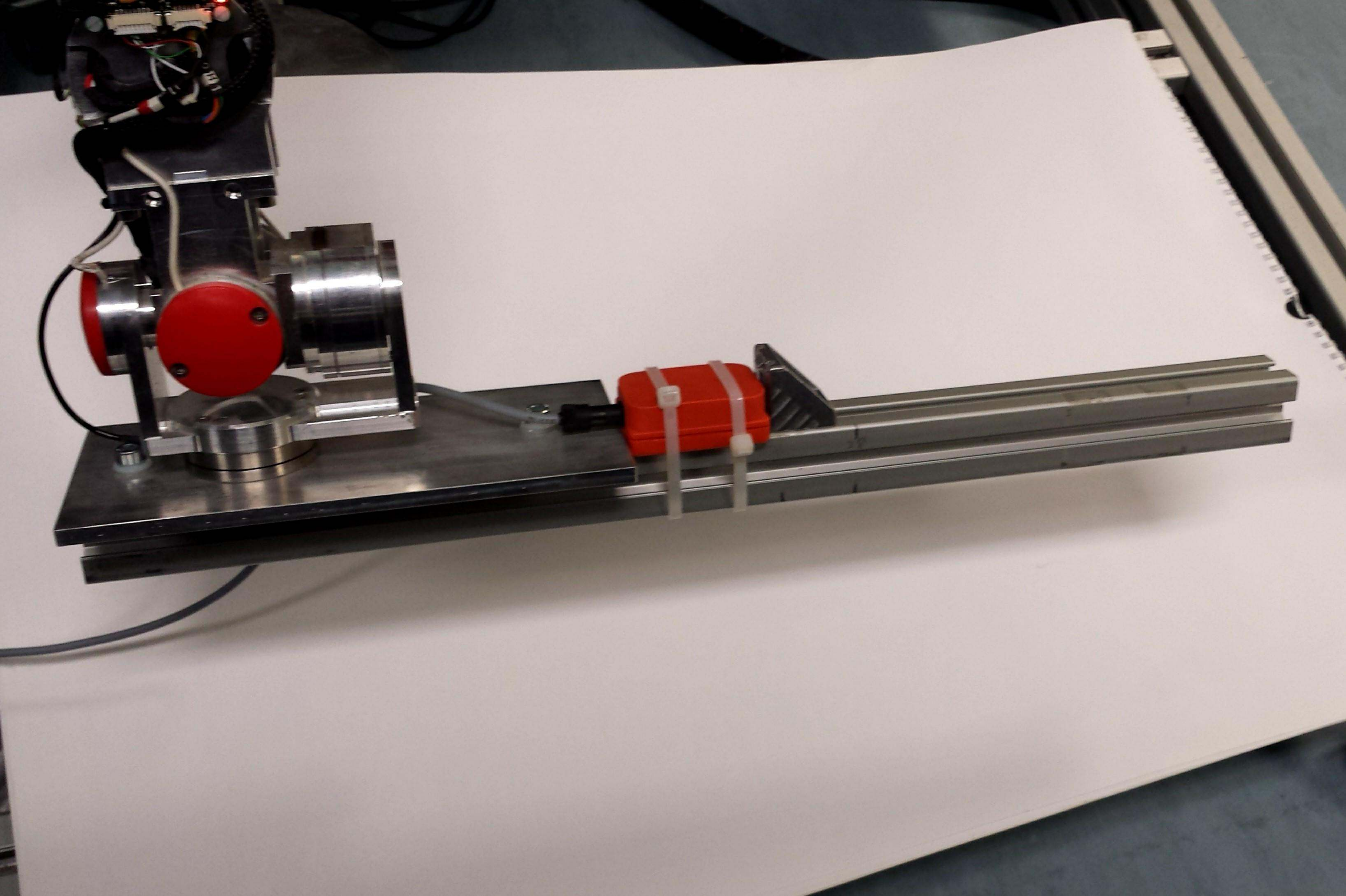}
\label{fig:dataset1}}
\subfloat[Dataset 2]{\includegraphics[width=0.22\textwidth]{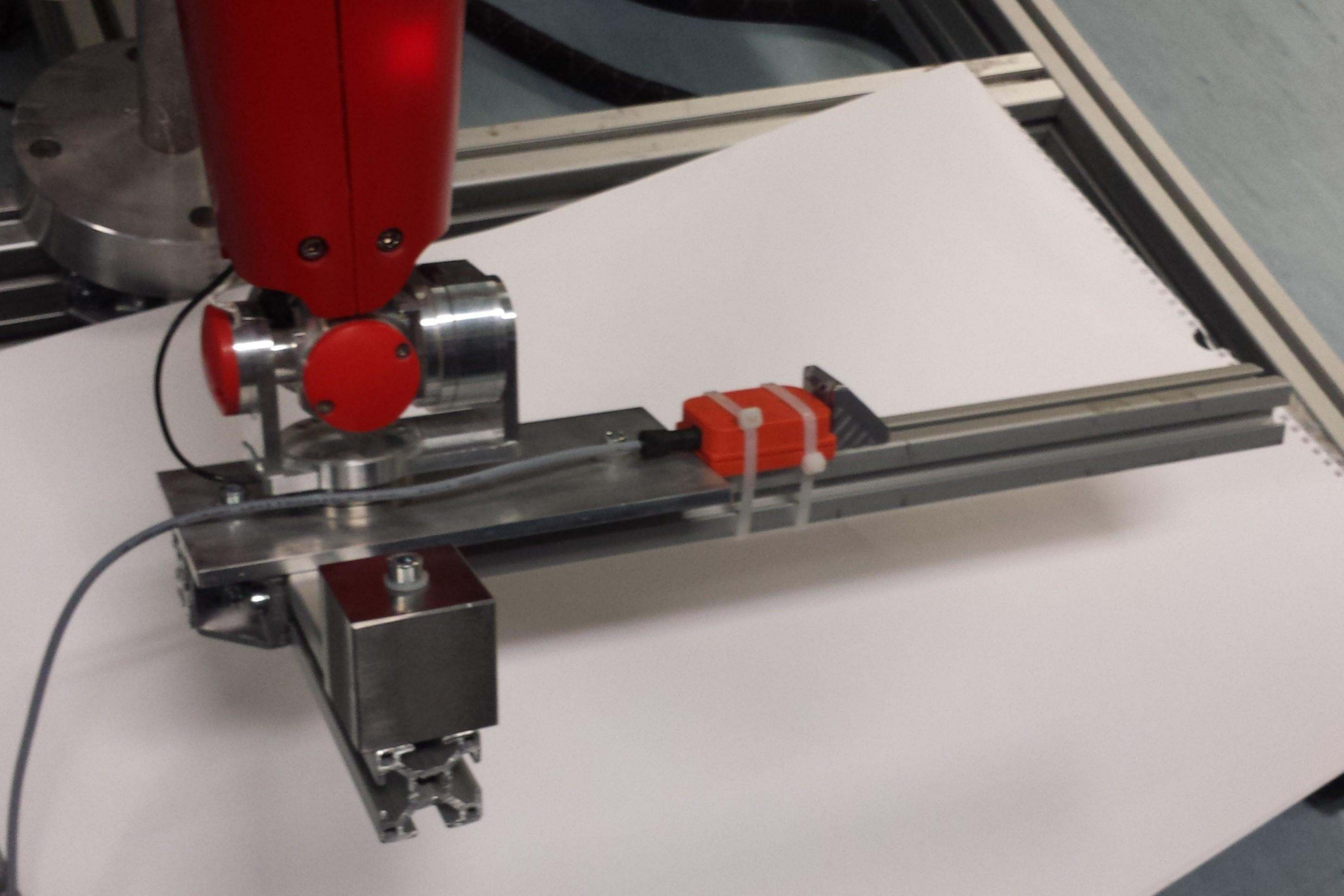}
\label{fig:dataset2}}
 \newline 
\subfloat[Dataset 3]{\includegraphics[width=0.22\textwidth]{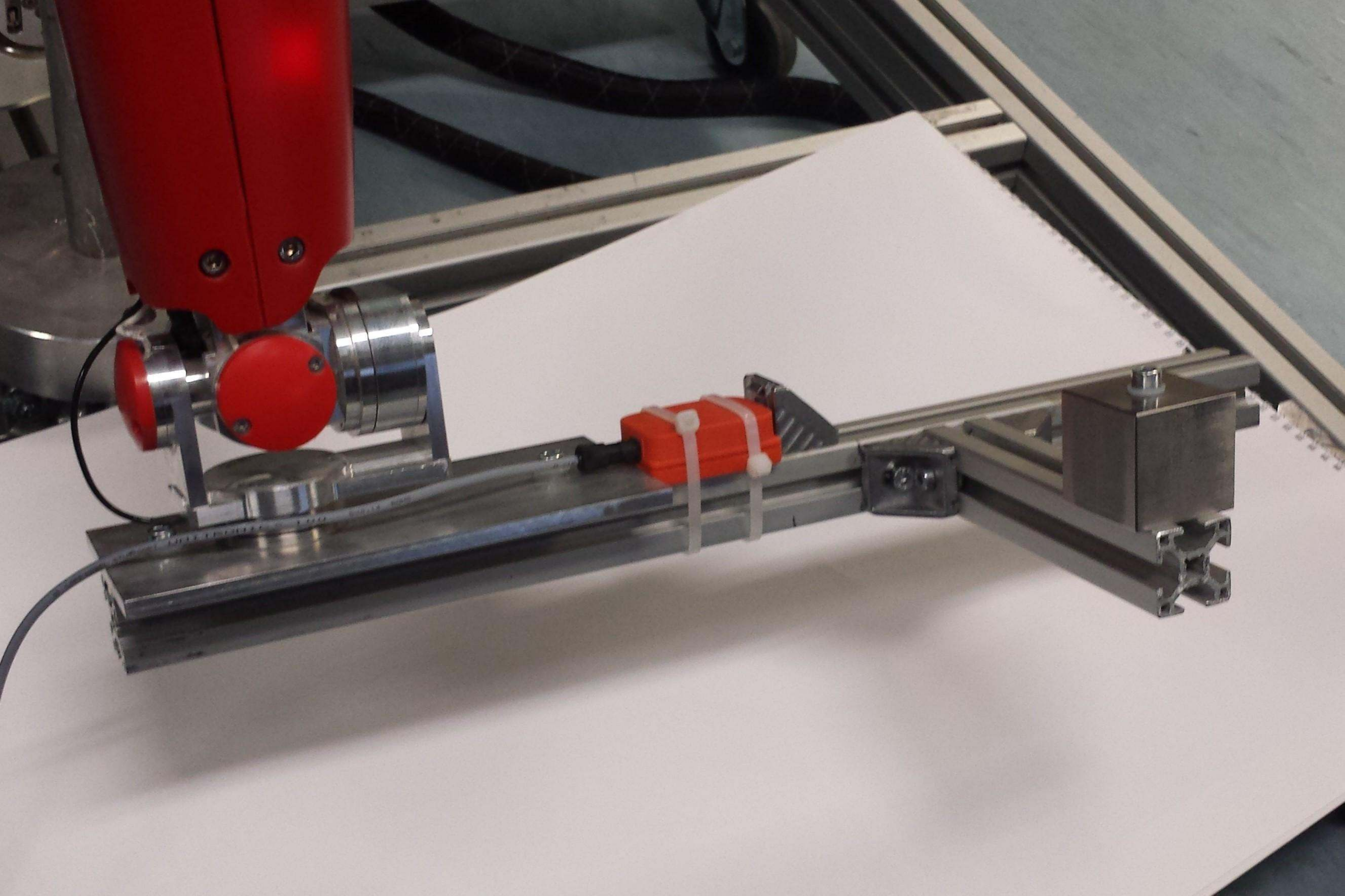}
\label{fig:dataset3}}
\subfloat[Dataset 4]{\includegraphics[width=0.22\textwidth]{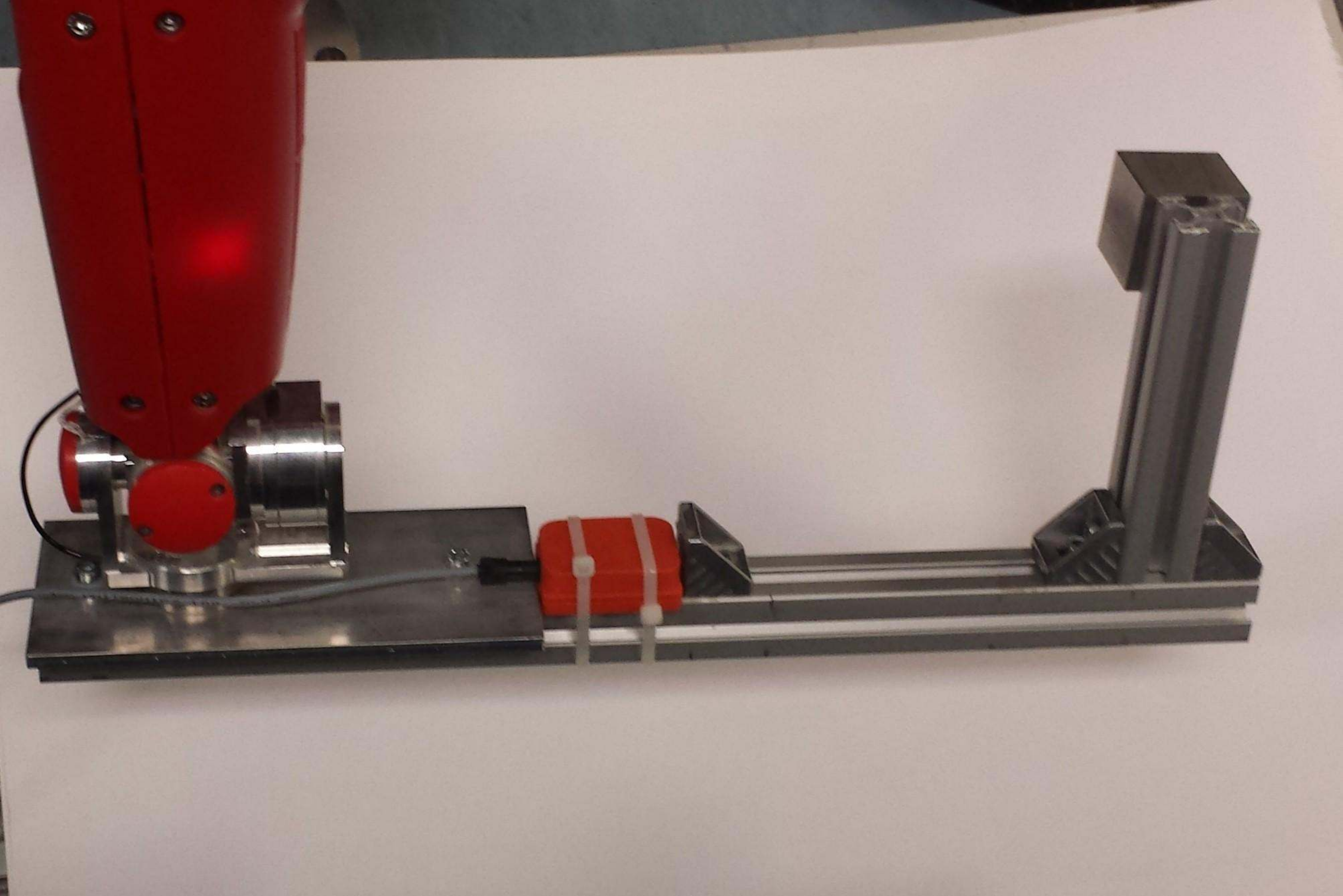}
\label{fig:dataset4}}
\caption{Added mass configurations for calibration datasets.}
\label{fig:calibration}
\end{figure}

\begin{proof}
This is a proof by contradiction. Assume $N_D~=~2$. In addition, assume, without loss of generality, also that the matrix $M_j$ in 
equation~\eqref{rawMeasurementsNoOffsetSeveralDataSets} 
is perfectly known (adding unknowns to the considered problem would only require a larger number of data sets). 
Then, in view of~\eqref{rawMeasurementsNoOffsetSeveralDataSets} and~\eqref{dataSets}, one has
\begin{IEEEeqnarray}{rcl}
 C(R_1,R_2) = (M_1G_1,M_2,G_2). \nonumber
\end{IEEEeqnarray}
The matrix $C$ is the unknown of the above equation. By applying \eqref{eq:kroneckerVec} one easily finds out that there exists a unique~$C$ only
if the rank of the matrix $(R_1,R_2)$ is equal to six, i.e. 
\[\text{rank}\Big( (R_1,R_2) \Big) = 6. \]
Recall that the matrix $C$ is invertible by assumption, and thus with rank equal to six. 
Consequently
\begin{IEEEeqnarray}{RCL}
 \text{rank}\Big( (R_1,R_2) \Big) &=& \text{rank}\Big( (M_1G_1,M_2,G_2)\Big) \nonumber \\
                                  &=& \text{rank}\left( (M_1,M_2)
                                  \begin{pmatrix}
                                    G_1 && 0 \\
                                    0 && G_2
                                  \end{pmatrix} \right) \nonumber \\
                                  &\leq& \min\Big(\text{rank}(M_1,M_2),6\Big).
				  \nonumber
\end{IEEEeqnarray} 
In view of~\eqref{matrixM}, one easily verifies that $\det(M_1,M_2) \equiv 0$, which implies that  \[\text{rank}\Big( (R_1,R_2) \Big) \leq 5.\] 
\end{proof}

Establishing a sufficient condition for the uniqueness of the solution~$x$ to the equation~\eqref{equationForEstimatingC}
is not as straightforward as proving the above necessary condition, and is beyond the scope of the present paper. Clearly, this uniqueness is related to 
the rank of the matrix $\Theta$, and this rank condition can be verified numerically on real data.
Then, the solution~$~x$ can be found by applying least-square techniques, thus yielding estimates of the calibration matrix~$C$ and of the inertial
characteristics of the rigid body.

\section{EXPERIMENTS}
\label{experiments}
To test the proposed method, we calibrated the two force-torque sensors embedded in the leg of the iCub humanoid robot 
-- see Figure~\ref{fig:iCubLeg}. The mass and the center of mass of this leg are unknown.

To apply the method described in section~\ref{method}, we need to add sample masses to the iCub's leg.
For this purpose, we installed on the robot's foot a beam to which samples masses can be easily attached. This beam also houses a XSens MTx IMU.
The supplementary accelerometer will no longer be required when using
 the iCub version 2, which will be equipped with 50 accelerometers 
distributed on the whole body. 
The iCub's knee is kept fixed with a position controller, so we can consider the robot's leg as a unique rigid body.
Consequently, the accelerometer measures the gravity force w.r.t. both sensors.  Recall also that 
to apply the proposed methods, we need to modify the orientations of the sensors' frames. To do so, we modified the \emph{front-back} and
\emph{lateral} angles associated with the robot's hip.

\begin{figure}[t]
\centering
\subfloat[Dataset 5]{\includegraphics[width=0.22\textwidth]{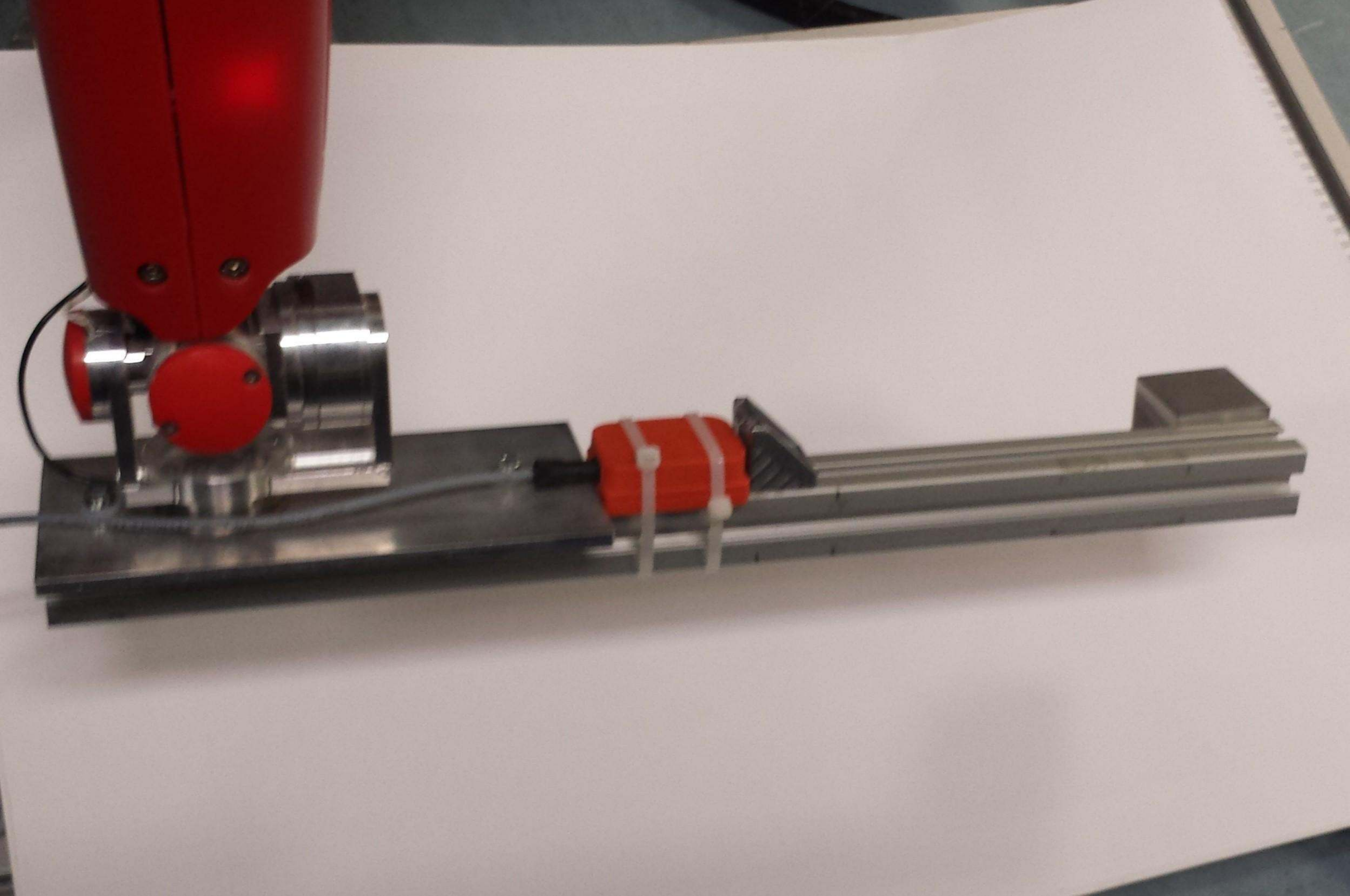}
\label{fig:dataset5}}
\subfloat[Dataset 6]{\includegraphics[width=0.22\textwidth]{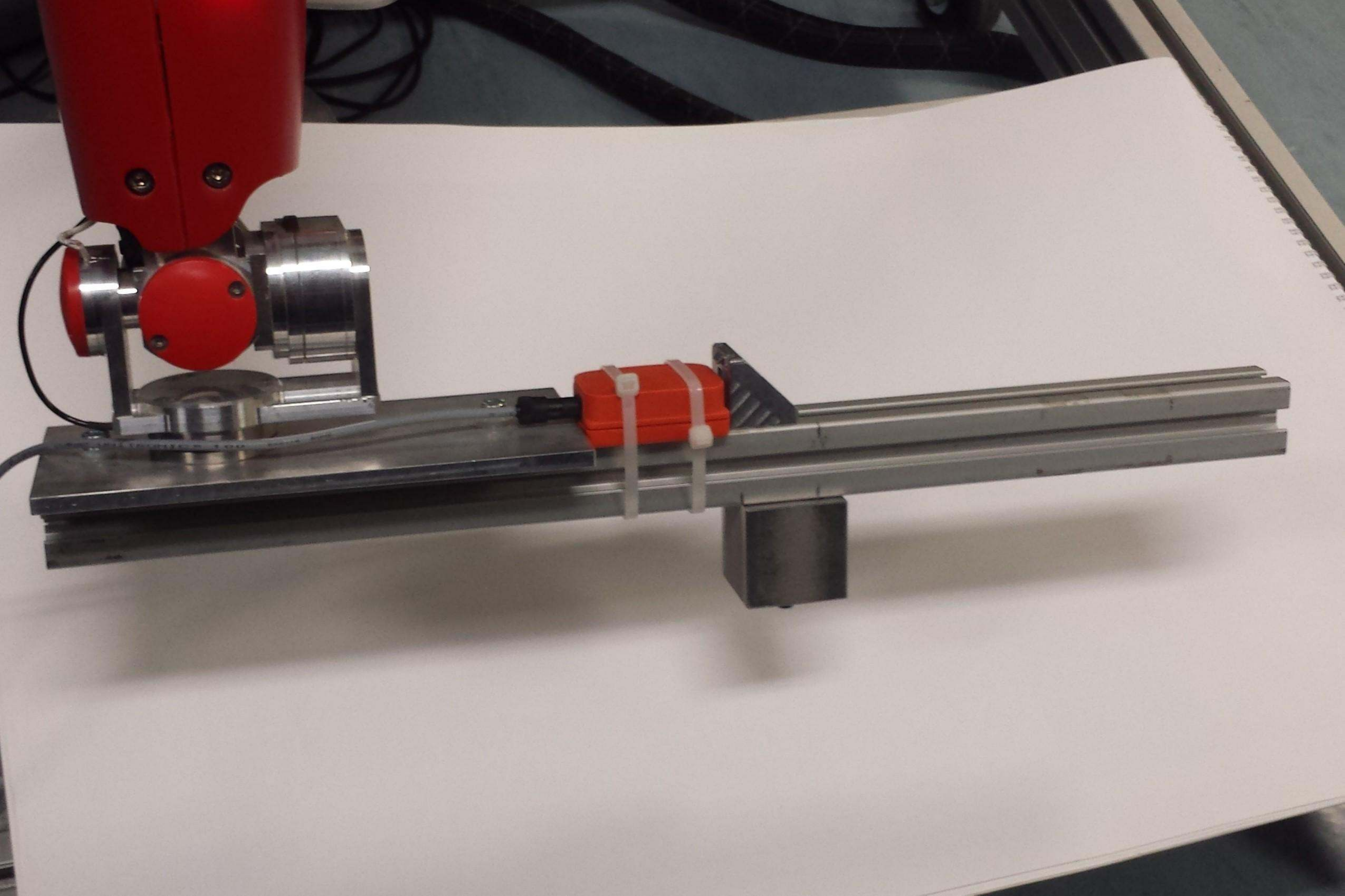}
\label{fig:dataset6}}
 \newline 
\subfloat[Dataset 7: no added mass]{\includegraphics[width=0.22\textwidth]{images/dataset0107.pdf}
\label{fig:dataset7}}
\subfloat[Dataset 8]{\includegraphics[width=0.22\textwidth]{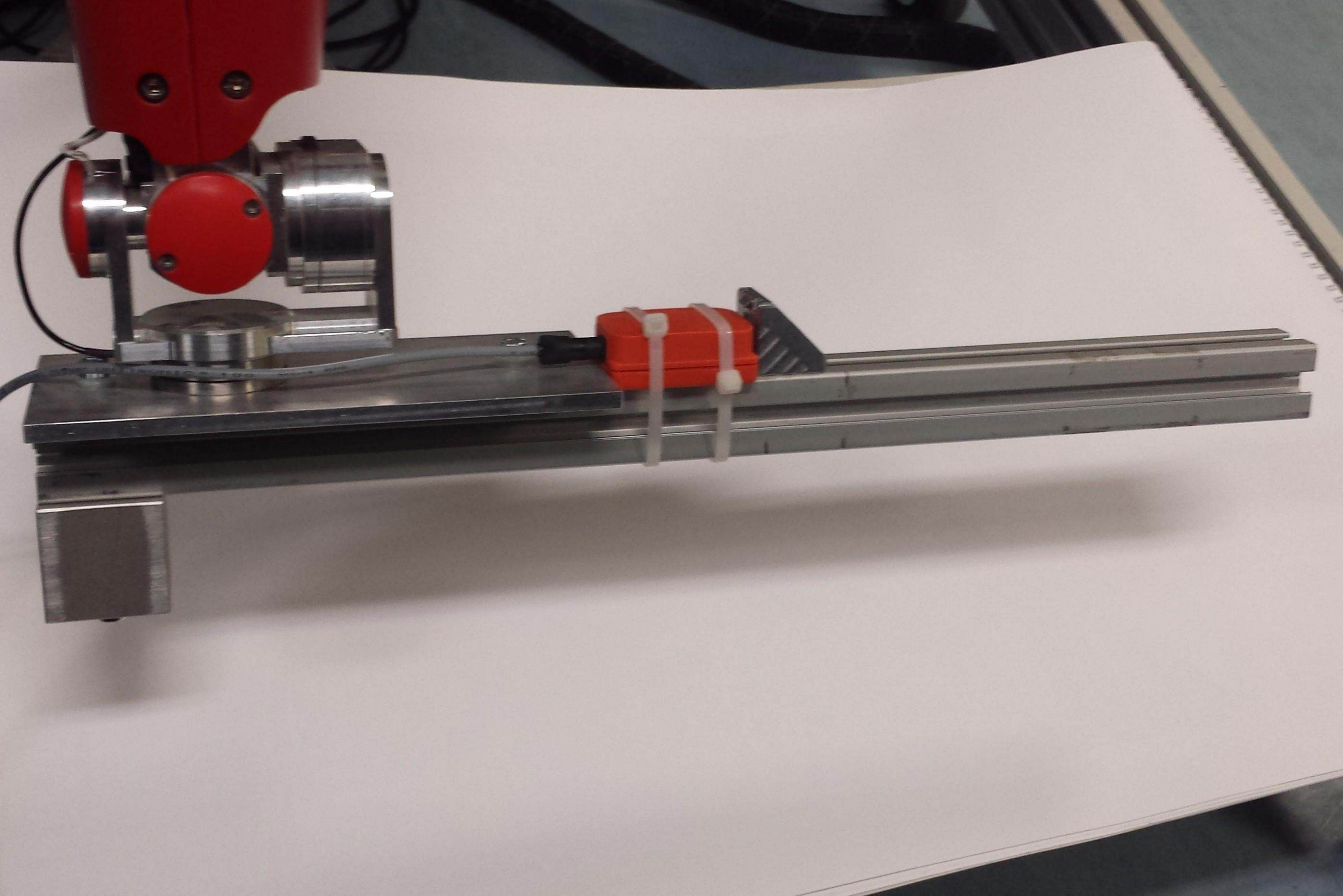}
\label{fig:dataset8}}
\caption{Added mass configurations for validation datasets.}
\label{fig:validation_masses}
\end{figure}

We collected data associated with eight different added mass configurations, each of which is characterized by a mass placed at a different 
location with respect to the beam. In other words, 
we collected eight different data sets. Figures~\ref{fig:calibration} and~\ref{fig:validation_masses} show the configurations of these data sets.

For each of these data sets, we slowly\footnote{The iCub front-back and lateral angles were moved with peak velocities of $2 \deg/s$.} 
moved the \emph{front-back} and
\emph{lateral} angles of the robot hip, spanning a range of $70 \deg$ for the \emph{front-back} angle, and a range of $90 \deg$ for the lateral angle. 
We sampled the two F/T sensors and the accelerometer at 100 Hz, and we filtered the obtained signals with a Savitzky-Golay filter of third order 
with a windows size of 301 samples. 

\begin{figure}
\vspace{0.5em}
 \centering
 \includegraphics[width=0.52\textwidth]{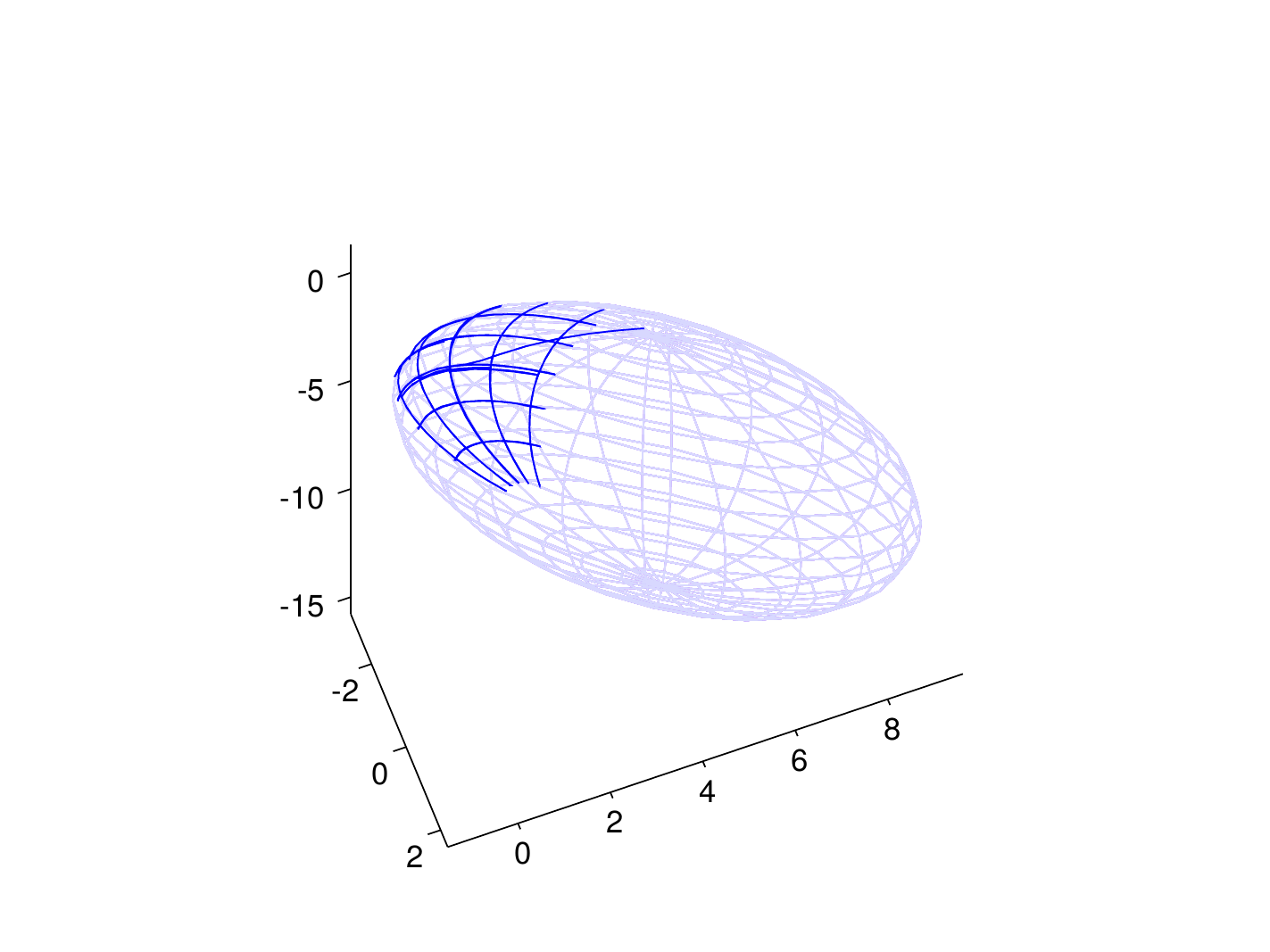}
 \caption{Dark blue: raw measurements of the sensor embedded in the leg for dataset 1 projected in the 3D subspace through $U_1$.
 In light blue an ellipsoid fitted to the measured points is added, to highlight the fact that the measured data lie on an ellipsoid. The $o'$ point estimated 
 with the method described in \ref{offsetEstimationTechnique} is the center of this ellipsoid.
  }
 \label{ellipsoidWithRawData}
\end{figure}

We estimated the sensors' offsets by applying the method described in section~\ref{offsetEstimationTechnique} on all eight data sets. 
Figure~\ref{ellipsoidWithRawData} verifies the statement of Lemma~1, i.e. the raw measurements belong to a three dimensional ellipsoid. 
In particular, this figure shows the measurements
$r_i$ projected onto the three dimensional space where the ellipsoid occurs, i.e. the left hand side of the equation~\eqref{rFromModel1}. 
Then, we removed the offset from the raw measurements to apply the estimation method for the calibration matrix described in 
section~\ref{calibrationMatrixEstimation}.

The two sensors' calibration matrices were identified by using only four data sets (see Figure~\ref{fig:calibration}). 
The other four were used to validate the obtained calibration results (see Figure~\ref{fig:validation_masses}).
The qualitative validation of the calibration procedure is based on the fact that the weight of the leg is constant for each data sets.
Consequently, if we plot
the force measured by the sensors, i.e. the first three rows of left hand side of the sensor's equation
\[ w = C(r - o), \]
these forces must belong to a sphere, since they represent the (constant norm) gravity force applied to the sensors.
As for elements of comparisons, we can also plot the first three rows of the above equation when evaluated with the calibration matrix that was 
originally provided by the
manufacturer of the sensors. 

\begin{figure}[h]
\vspace{0.5em}
\centering
\subfloat[Validation results for leg F/T sensor.]{\includegraphics[width=0.45\textwidth]{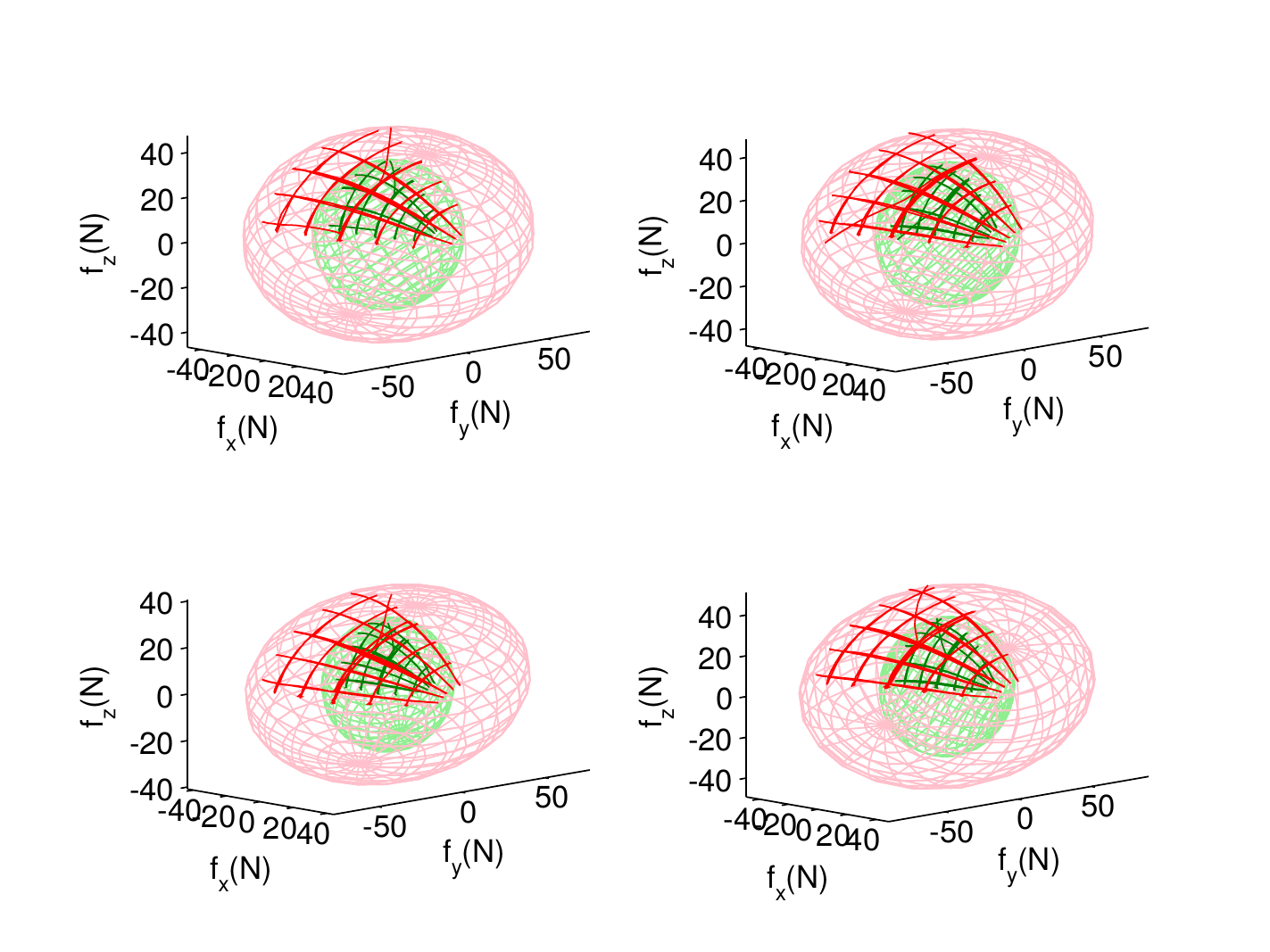}}
\newline
\subfloat[Validation results for foot F/T sensor.]{\includegraphics[width=0.45\textwidth]{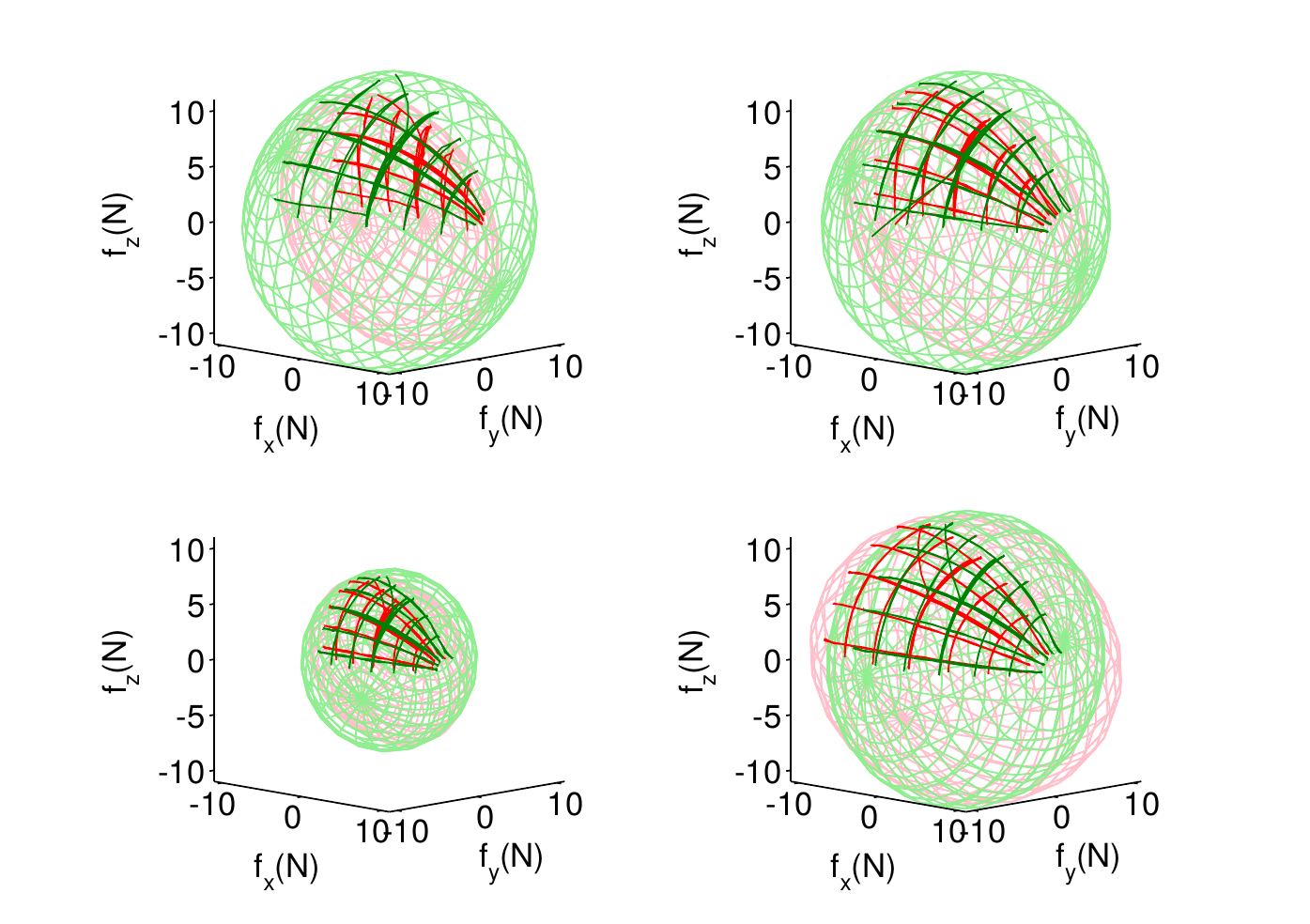}}
\caption{Dark green: force measurements obtained through the calibration matrix estimated using the proposed technique. Dark red:
force measurements obtained through the calibration matrix provided with the sensor. Light red and light green surfaces: ellipsoids fitted to the 
measured forces.}
\label{fig:validation}
\end{figure}

Figure~\ref{fig:validation} depicts the force measured by the sensor with the estimated calibration matrix (in green)
and with the calibration matrix provided by the manufacturer (in red). It is clear to see that the green surfaces are much more spherical than the red ones.
As a matter of fact, Table \ref{table:soa} lists the semi axes of the ellipsoids plotted in Figures~\ref{fig:validation}, and clearly shows 
that the green surfaces represent spheres much better than the red ones. 
Interestingly enough, the force-torque sensor embedded in the iCub leg is much more mis-calibrated than that embedded in 
the foot. In fact, by looking at the data sheets describing the technological lives of these sensors, we found out that the force-torque sensor embedded in
the leg is much older than that in the foot, which means that the calibration matrix of the leg's sensor is much older than that of the foot's sensor.

The quantitative validation of the proposed calibration procedure is performed by comparing the known weights of the added masses with the weights estimated by the sensors. 
Table \ref{table:mass} shows that the estimated weights obtained after performing the proposed calibration are better than those estimated by using the calibration matrix
provided by the sensor manufacturer. A similar comparison was conducted on the estimation of the sample mass positions (Table \ref{table:cmass}). It has to be noticed that the errors in estimating the mass position are relatively high, but this is due to the choice of using relatively small masses with respect to the sensor range and signal to noise ratio.

\begin{table*}[ht] 

\caption{Qualitative calibration evaluation on validation dataset: ellipse semiaxes after calibration}
\centering 
\begin{tabular}{p{0.8cm} | p{1.2cm} | p{1.2cm}  | p{1.3cm} p{1.3cm} p{1.3cm} | p{1.3cm} p{1.3cm} p{1.3cm}} 
              \emph{Sensor}   & \emph{Dataset} & \emph{Added mass} (Kg) & \multicolumn{3}{p{3.9cm} |}{\emph{Semiaxes length [N] with proposed calibration}} & \multicolumn{3}{p{3.9cm}}{\emph{Semiaxes length [N] with manufacturer calibration}}\\
\hline \rowcolor[gray]{.9}
  \rowcolor[gray]{.9} Foot    &  Dataset 5     &  0.51  & 13.6 &  13.1 & 12.9 & 13.5 & 10.2 & 4.6  \\ 
  \rowcolor[gray]{.9} Foot	&  Dataset 6   &  0.51   & 13.5 & 12.9 & 12.7  & 13.6 & 10.5 & 9.4  \\
  \rowcolor[gray]{.9} Foot	&  Dataset 7   &  0   & 8.4 & 7.9 & 7.4 & 8.5 &  6.9  & 6.2   \\
  \rowcolor[gray]{.9}  Foot	&  Dataset 8   &  0.51   & 13.7 & 12.5 & 12.0 & 15.7 & 13.6 & 10.4  \\
 \hline 
 Leg  &  Dataset 5    & 0.51   & 34.4 & 33.3 & 32.5 & 76.5 & 49.4 & 45.6 \\ 
 Leg	&  Dataset 6  & 0.51  & 34.8 & 33.5 & 32.8  & 82.4 & 49.3 & 47.3  \\
 Leg	&  Dataset 7  & 0     & 30.9 & 28.2 & 26.9   & 77.0 & 44.5  & 40.0 \\
 Leg	&  Dataset 8  & 0.51   & 35.52 & 33.30 & 32.2 & 88.9 & 49.5 & 48.3   \\
\hline

\end{tabular} 
\label{table:soa} 
\end{table*}

\begin{table*}[ht] 
\caption{Qualitative calibration evaluation on validation dataset: sample mass estimations}
\centering 
\begin{tabular}{p{0.8cm} | p{1.2cm} | p{3cm}  | p{3cm} | p{4cm} } 
              \emph{Sensor}   & \emph{Dataset} & \multicolumn{3}{c}{\emph{Added mass} (Kg)} \\
\hline \rowcolor[gray]{.9}
\hline \rowcolor[gray]{.9}    &                &  Ground truth & With proposed calibration & With manufacturer calibration      \\
  \rowcolor[gray]{.9} Foot    &  Dataset 5     &  0.51    &  0.53  & 0.06  \\ 
  \rowcolor[gray]{.9} Foot	&  Dataset 6   &  0.51    &  0.52   &  0.27  \\
  \rowcolor[gray]{.9} Foot	&  Dataset 7   &  0       & -0.03  &  -0.13 \\
  \rowcolor[gray]{.9}  Foot	&  Dataset 8   &  0.51    &  0.46  & 0.45 \\
 \hline 
 Leg  &  Dataset 5    & 0.51  & 0.51 & 2.77  \\ 
 Leg	&  Dataset 6  & 0.51  & 0.54 & 3.04  \\
 Leg	&  Dataset 7  & 0     & -0.04  & 2.39  \\
 Leg	&  Dataset 8  & 0.51  & 0.51  & 3.25  \\
 \hline
 \end{tabular} 
 \label{table:mass} 
 \end{table*}

 \begin{table*}[ht] 
 \caption{Qualitative calibration evaluation on validation dataset: center of mass estimations}
 \centering 
 \begin{tabular}{p{1.8cm} | p{2.2cm} | p{1.4cm} |  p{0.7cm} |  p{0.7cm}  ||  p{0.7cm} |  p{0.7cm} |  p{0.7cm}||  p{0.7cm} |  p{0.7cm} |  p{1.2cm} } 
              \emph{Sensor}   & \emph{Dataset} & \multicolumn{9}{c}{\emph{Center of mass position for the added mass} [cm]} \\
 \hline \rowcolor[gray]{.9} -   &         -     &  \multicolumn{3}{c || }{Ground truth} & \multicolumn{3}{c ||}{With proposed calibration} & \multicolumn{3}{c}{With manufacturer calibration} \\ 
  \rowcolor[gray]{.9} Foot    &  Dataset 5     & 39  & {-}3.5 & 2.9  & 31  & 8.8 & $-$8.3  & 273  & $-$83 & 81  \\ 
  \rowcolor[gray]{.9} Foot	&  Dataset 6   &  21 &  0   & 6.3  & 19  & 9.9 & $-$5    & 30   & $-$18 & 18  \\
  \rowcolor[gray]{.9} Foot	&  Dataset 7   &  -  & -    & -    & -   &  -  & -     & -    &  -  & -  \\
  \rowcolor[gray]{.9}  Foot	&  Dataset 8   &  {-}4 & 0   & 6.3  & 5.5 &  9  & $-$3.2  & $-$8.6 & $-$12 & 10 \\
 \hline 
 Leg  &  Dataset 5    & 39 & $-$3.5 & 39  & 28 & 11 & 29 & 16 & 6.5 & 29  \\ 
Leg	&  Dataset 6  & 20 & 0  & 43  & 17 & 9.7 & 30 & 12 & 5.1 & $-$27  \\
Leg	&  Dataset 7  & -  & -  & -   & -  & -   & -  & -  & -   & -  \\
 Leg	&  Dataset 8  & $-$4.3 & 0 & 43 & 3.8 & 8.8 & 32 & 7.5 & 4.9 & $-$26 \\
\hline
\end{tabular} 
\label{table:cmass} 
\end{table*}


\section{CONCLUSIONS}
\label{conclusions}
In this paper, we addressed the problem of calibrating six-axis force-torque sensors in situ by using the accelerometer measurements. 
The main point was to highlight the geometry behind the gravitational
raw measurements of the sensor, which can be shown to belong to a three-dimensional affine space, and more precisely to a three-dimensional ellipsoid.
Then, we propose a method to identify first the sensor's offset, and then the sensor's calibration matrix. The latter method requires to add sample masses
to the body attached to the sensor, but is independent from the mass and the center of mass of this body. We show that a necessary condition to identify the sensor's calibration matrix is to collect data for more than two 
sample masses. The validation of the method was performed by calibrating the two force-torque sensors embedded in the iCub leg.

The main assumption under the proposed algorithm is that the measurements were taken for static configurations of the rigid body attached to the sensor.
Then, future work consists in weakening this assumption, and developing calibration procedures that hold even for dynamic motions of the rigid body. This
extension requires to use the gyros measurements. In addition, comparisons of the proposed method versus existing calibration 
techniques is currently being investigated, and will be presented in a forthcoming publication.



\bibliographystyle{IEEEtran}
\bibliography{bibl}

\end{document}